\documentclass[11pt,a4paper]{article}

\usepackage{acl2014}
\usepackage{times}
\usepackage{latexsym}
\usepackage{amsmath}
\usepackage{color}
\usepackage{epsfig,url,algorithm,algorithmic,multirow}
\usepackage{amssymb}

\newcommand{\squishlist}{
 \begin{list}{$\bullet$}
  { \setlength{\itemsep}{0pt}
     \setlength{\parsep}{3pt}
     \setlength{\topsep}{3pt}
     \setlength{\partopsep}{0pt}
     \setlength{\leftmargin}{1.5em}
     \setlength{\labelwidth}{1em}
     \setlength{\labelsep}{0.5em} } }

\newcommand{\squishlisttwo}{
 \begin{list}{$\bullet$}
  { \setlength{\itemsep}{0pt}
     \setlength{\parsep}{0pt}
    \setlength{\topsep}{0pt}
    \setlength{\partopsep}{0pt}
    \setlength{\leftmargin}{2em}
    \setlength{\labelwidth}{1.5em}
    \setlength{\labelsep}{0.5em} } }

\newcommand{\squishend}{\end{list}}

\newcommand{\ignore}[1]{}
\newcommand{\comment}[1]{}

\renewcommand{\paragraph}[1]{\noindent\textbf{#1.}}

\usepackage{times}
\usepackage{latexsym}
\usepackage{amsmath}
\usepackage{bbm}
\usepackage{multirow}
\usepackage{url}

\usepackage{booktabs}
\usepackage{graphicx}
\usepackage{varwidth}

\title{Bootstrapping Ternary Relation Extractors}

\author{Ndapandula Nakashole \\
  Carnegie Mellon University \\
  5000 Forbes Avenue \\
  Pittsburgh, PA, 15213 \\
  {\tt ndapa@cs.cmu.edu} \\}

\date{}

\begin{document}

\maketitle

\begin{abstract}
Binary relation extraction methods  have been widely studied in recent years.  
However, few methods have been developed for higher n-ary relation extraction. 
One  limiting factor is the effort required to generate training data.
  For binary relations, one only has  to provide a few dozen  pairs of entities per relation, as training data.  For ternary  relations (n=3), each training instance is a  triplet of entities, placing  a greater  cognitive load on people.  For example, many people know that Google acquired Youtube but not the dollar amount or the date of the acquisition and many people know that Hillary Clinton is married to Bill Clinton by not the location or date of their wedding. This makes higher n-nary training data generation  a  time consuming exercise in searching the Web. We present  a resource for training ternary relation extractors.  This was  generated using a minimally supervised yet effective approach. We present statistics on the size and  the quality of the dataset.

\end{abstract}

\section{Introduction}
%

Developing techniques for higher-nary relation extraction  is a natural next step  after  the well-studied case of binary relations \cite{Auer07,suchanek2007yago,Bollacker2008,Carlson2010,MitchellCHTBCMG15}.  
  In the literature, prominent  binary relation extraction methods  are mostly  semi-supervised \cite{Suchanek:2009,Carlson2010} or  unsupervised  \cite{Mausam2012,fader2011identifying}. Semi-supervised methods tend to have higher precision than unsupervised methods and therefore are commonly used to populate knowledge bases  of facts \cite{Nakashole2011,MitchellCHTBCMG15}. In such settings,  relations of interest are predefined, i.e, company acquisitions or  protein-protein interactions.
However,  in semi-supervised approaches,  one needs to provide seed examples for each relation to bootstrap the extractor. This can be expensive, especially if there are many relations of interest. For ternary relations, hand specifying training instances per relation  requires even more time since each training instance is a triplet of three entities as opposed to a pair of entities for binary relations.  Most people know that Google acquired Youtube but not the dollar amount or the date of the acquisition, and most people know that Hillary Clinton is married to Bill Clinton by not the location or date of their wedding. This makes ternary training data generation  a  time consuming exercise in searching the Web. 
 
In this paper we present a resource  for training  ternary extractors. The resource was generated using a minimally supervised yet  high precision method.
Our method leverages a very common language construction: prepositional phrases (PPs).   PPs such as  ``in X", ``at Y", and ``for Z" express details about the  \textit{where, when,} and \textit{why}  of  binary relation instances. This makes PPs well-suited to extending binary relations by one more argument, extending them to ternary relations.
 
 Consider the following occurrences of 5-item sequences of the form: N1, V, N2, P, N3; where the Ns are noun phrases,  V is a verb and P is a preposition.
 \begin{table}[h]
    \small{
 \centering
 \begin{tabular}{p{7cm}}
 \textit{(1) \textbf{Mercedes-Benz}	bought	\textbf{Chrysler}	for	\textbf{\$40 billion}}\\
   \textit{(2) \textbf{CBS}	bought	\textbf{WCCO}	from	\textbf{General Mills}}\\
  \hline
  \textit{(3) \textbf{Joe Lieberman}	endorsed	\textbf{McCain}	for	\textbf{president}}\\
  \textit{(4) \textbf{The New Yorker}	endorsed	\textbf{Obama}	over	 \textbf{Romney}}\\
 \end{tabular}
 \label{tbl:example}
 }
 \end{table}

In a large Web extracted corpus, one sees many 5-item sequences similar to each of the 6 types shown above. The verbs and prepositions do not change but the arguments N1-3 change. From a large volume of such occurrences, we can learn templates for populating ternary relations. For example, from many occurrences of tuples of type (1), we can generate the template:   \textit{ \textless organization\textgreater	bought \textless organization\textgreater	for	\textless dollar\_amount\textgreater}. We go further by having labels for all three argument placeholders.  For this particular template the labels are: \textit{AcquisationEventAcquirer, AcquisationEventAcquired, AcquisationEventAmount} for the N0, N1, N3, argument placeholders respectively.
Similarly, from tuples of type (3), we can generate the template:   \textit{ \textless person\textgreater	endorsed \textless politician\textgreater	for	\textless political\_office\textgreater}. And the corresponding argument labels are: \textit{EndorsementEventEndorser,	EndorsementEventEndorsed,	EndorsementEventOffice}, for the N0, N1, N3 argument placeholders respectively. A triplet of argument labels is considered to be a ternary relation. And matching triplets of entities are considered to be instances of ternary relations.

In summary, the contributions of this paper are twofold:

\textit{1)~ Resource:}  A resource for boostrapping ternary relation extraction. It contains ternary relations and their instances. It also contains templates used to populate ternary relations. We make the data available for future research. It is also attached as supplementary data to this submission.

\textit{2)~ Data Generation Method:} We describe  the approach used to generate this resource, which others can replicate to generate similar resources in their  domains of interest.

  \section{Related Work}
  To supervise binary relation extraction methods,  there is an abundance  of resources. Knowledge bases (KBs) such as  DBPedia \cite{Auer07}, YAGO \cite{suchanek2007yago}, Freebase\cite{Bollacker2008}, and NELL \cite{MitchellCHTBCMG15}, as well as Open IE tools and resources such as Reverb \cite{Fader2011}, and Ollie \cite{Mausam2012} contains many millions of binary relation instances that can be used  to distantly train binary relation extraction. However, these knowledge bases are highly impoverished when it comes to ternary relations.  In contrast, we provide a resource that can be used directly to train and encourage more research on the higher-nary machine reading and relation extraction. A number of works have studied temporal scoping of  facts by adding a time dimension to facts \cite{wang2010timely,wijayactp}. While temporal scope generates ternary relations, for example by using reification, this only deals with one type of ternary relation. In contrast, our resource contains $50$ ternary relations, spanning $18$ high level topics or  event types.
  
  \begin{table}[t]
  \begin{tabular}{|l|l|}
  \hline
  ElectionEvent & AwardEvent \\
  HiringEvent & FiringEvent \\
  AcquisitionEvent & WeddingEvent \\
  DivorceEvent & DefeatEvent \\
  MeetingEvent & AttackEvent \\
  ProductLaunchEvent & EarthquakeEvent \\
  MurderEvent & PerformingEvent \\
  SuingEvent & BombingEvent \\
  EndorsementEvent & ShootingEvent \\
  \hline
  \end{tabular}
  \caption{Event types (18) or high level topics in our resource.  The resource contains 50 ternary relations across these topics.}
  \label{eventtypes}
  \end{table}
  
 \section{Data Generation}
In this section we present our method for generating the ternary relations and their instances.

\subsection{Input}
As input, our method takes a natural language text corpus and   high level topics or even types of interest. Our method automatically learns many different ternary relations relevant for each event type. There are two advantages to  specifying event types of interest instead of directly thinking in terms of  ternary relations.  First, a broad event type can capture many relevant ternary relations that naturally appear in the data. Second, it requires much less human effort to specify one event type than manually specifying a list of all conceivable  ternary relations, some of which might not be present in the data.

Table \ref{eventtypes} shows the event types covered in our resource. For each event type, we specified at most three \textit{trigger verbs} that indicate a potential mention of the event type. We will later describe how to automatically extend  event trigger verbs in an iterative manner. This is done in a similar way to extending seed instances or patterns of  binary relations \cite{Suchanek:2009,Nakashole2011,MitchellCHTBCMG15}.

\subsection{Candidate Template Generation} \label{candidates}
Once we have event types defined with their trigger verbs, we can  generate ternary relations
for each event type. The first step is to generate ternary relation \textbf{templates}. An example template is: \textit{ \textless person\textgreater	endorsed \textless politician\textgreater	for	\textless political\_office\textgreater}.  We generate these templates directly from the data. We do this by first parsing the raw corpus, and extracting 5-item sequences of the form:  N1, V, N2, P, N3; where the Ns are noun phrases,  V is a verb such that V is a trigger verb of one of the events,  and P is a preposition.
We generate templates from 5-item sequences as follows: we replace every noun phrase N1-N3 by its semantic type.
In particular, we do lookups of entity types in two types of semantic hierarchies, WordNet and the NELL type hierarchy \cite{Carlson2010}. We found the two type systems to be complementary: WordNet contains more common nouns whereas NELL contains more proper nouns.
Our generated templates can therefore contain a mixture of WordNet types and NELL types. For example, for the \textit{MurderEvent}, the following is a valid template that our approach generated: \textit{\textless NEL\_person\textgreater	killed	\textless NEL\_person\textgreater	with	\textless WDN\_weapon\textgreater}. The semantic types in our templates are prefixed by three letters, \textit{NEL} for NELL types, and \textit{WDN} for WordNet types.

We retain, as candidate templates, all the templates of the form:
\textit{\textless N1\_type\textgreater	V	\textless N2\_type\textgreater	P	\textless N3\_type\textgreater}, whose support size is $>=3$. That is, the template was generated from three or more  5-item sequences, N1, V, N2, P, N3, with distinct noun arguments (N1-N3).

\subsection{Template Filtering}\label{filteringandinstances}
From the candidate templates, a final set of templates is generated.
To do this, we manually filter out all templates that do not express useful
ternary relations for the topic at hand. Once we have filtered out the templates, for each
template, we manually label each template with  descriptive labels for the its corresponding 
noun phrase placeholders. For example, \textit{\textless NEL\_person\textgreater	killed	\textless NEL\_person\textgreater	with	\textless WDN\_weapon\textgreater} is labeled with \textit{MurderEventMurderer, MurderEventMurdered, MurderEventInstrument}. Each triple of labels is considered to  be a single \textbf{ternary relation}. Therefore we have the ternary relation \textit{Murderer\_Murdered\_Instrument}. Such a ternary relation would  has \textbf{instances} such as \textit{Bob,Alice,knife},  which indicates that Bob murdered Alice with a knife.

We obtain instances of templates, and hence ternary relations, by  retaining the supporting 5-item sequences of each of the accepted templates.
 Notice that each instance  has labeled noun phrases  because the instances inherit  argument labels from their templates.

It is worth noting that  template filtering is the part requiring the most manual supervision. All the other parts are automated.  While, the specification of event types and their trigger verbs is also manual, it is quite fast, requiring only up to three trigger verbs per event type.

 
 \subsection{Iterative Template Generation} \label{iterate}
 In order to extend the size of the resource, we  increase the number of trigger verbs per event type. We  do this automatically. First, from the raw data, we extract 5-item sequences of the form $N1, V, N2, AP, N3$. There are two main differences between these sequences and the 5-item sequences we have worked with up to now. First,  here $V$ is any verb, not limited to the trigger verbs that were manually specified. Second, we no longer limit the phrase between $N2$ and $N3$ to prepositional phrases, $P$ is now $AP=any$ $phrase$. We limit the length of $AP$ to a maximum of three words.  
 We then find 5-item sequences where all three arguments $N1-3$ match the arguments of an instance of a template from Section~\ref{filteringandinstances}.  Thus, we are using the instances generated so far as distant supervision to discover new templates. 
 
 All new $(V,AP)$ pairs forming candidate templates that occur with more than $10$ distinct instances of an existing template qualify  a new promoted  templates. We increase the minimum support required from $3$ to $10$ to avoid introducing noisy templates.
A new template has  the same argument role labels as the original template whose instances overlap with the instances of the new template.  This process extends the trigger verbs, and is not  limited to prepositions for the extraction of the third argument.   This also allows the generation of more instances from the newly discovered templates. Notice that the number of ternary relations  remain constant from the initial template generation step where we manually label templates with argument roles.

%

\section{Evaluation}
We applied our data generation process to  Wikipedia (WKP) and the  ClueWeb09 (CWB) corpus. 
We first generated an initial set of candidate  templates from Wikipedia, using the method described in Section \ref{candidates}. We did not apply this step to the ClueWeb corpus as it can be noisy. We then manually filtered the generated candidate templates as described in Section \ref{filteringandinstances}. This is  iteration $0$. We had a total of $186$ templates at iteration $0$, and  $50$ ternary relations across $18$ event types.  The number of templates increase across iterations.
 Therefore, in subsequent iterations, we obtain more templates that can be used to populate our $50$ ternary relations. 

 From the  initial templates, we generate template instances both from Wikipedia and the ClueWeb corpus. These are triples of entities whose types match the argument types of the template, and they occur with the lexical items that appear in the template. We then use the instances as distant supervision to generate more templates as described in Section \ref{iterate}. Again, we only discover new templates from Wikipedia, using only ClueWeb to find instances for the discovered templates. At iteration 1, we discovered an additional 174 templates.
 
\begin{figure}[t]
 \centering
 \includegraphics[width=0.80\columnwidth] {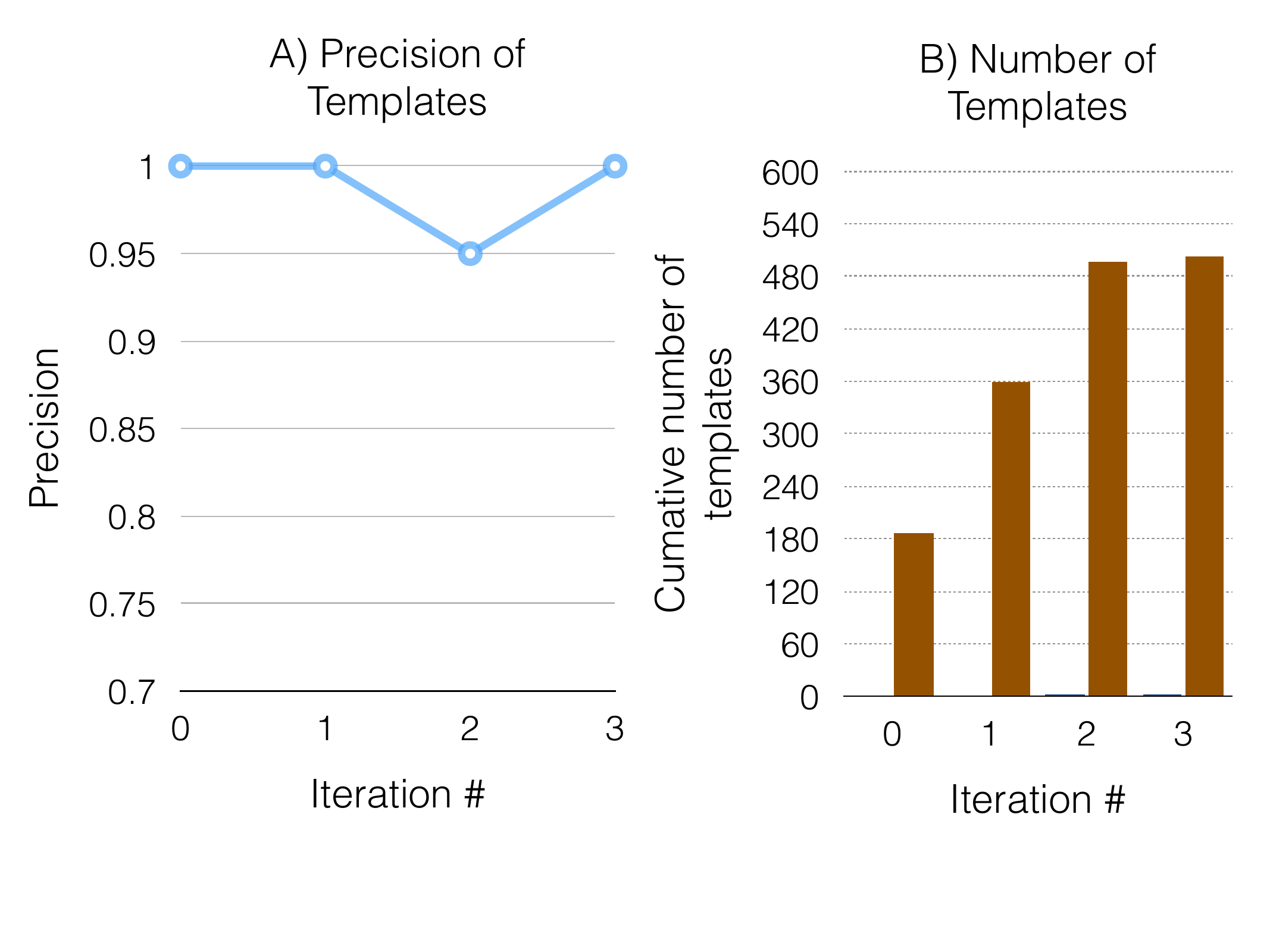}
 \vspace*{-0.98cm}
 \caption{Precision and accumulated number of learned templates, iterations $0-3$}
 \label{fig:templates}
 \end{figure}

 Our method picked up new templates until iteration 3, making a total of $502$ templates. Figure \ref{fig:templates} shows the cumulative number of templates across iterations. Also shown in Figure \ref{fig:templates}  is precision of templates.  We manually assessed precision at every iteration, we did this by randomly selecting $100$ templates discovered at every iteration, or all templates if less than $100$ templates were discovered in a given iteration. Precision is $100$\%  at all iterations, except for iteration 2 where it dropped to $95$\%. 
This was due to a   few cases of semantic drift not being cutoff by thresholds of our methods. This led our method do discover templates with verbs such as ``resigned as" in templates associated with \textit{hiring} events, we marked these as wrong.

Figure \ref{candidates} shows the cumulative number of instances picked up at every iteration. We started with $31,161$ from WKP and CWB corpora at iteration $0$ and ended up with $61,380$ instances  by the third and final iteration. This number could be increased by 1) allowing discovery of templates from the ClueWeb corpus or other large corpora 2) lowering the high thresholds on the minimum support size of learned templates .
Since  instances are generated from templates, their precision can be   inferred from the precision of the templates. To a small extent, instances can also contains errors  stemming  from:  noun phrase chunking and  semantic types; these  errors can be fixed by using  accurate better chunkers  and semantic typing systems.

 \begin{figure}[t]
  \centering
  \includegraphics[width=0.80\columnwidth] {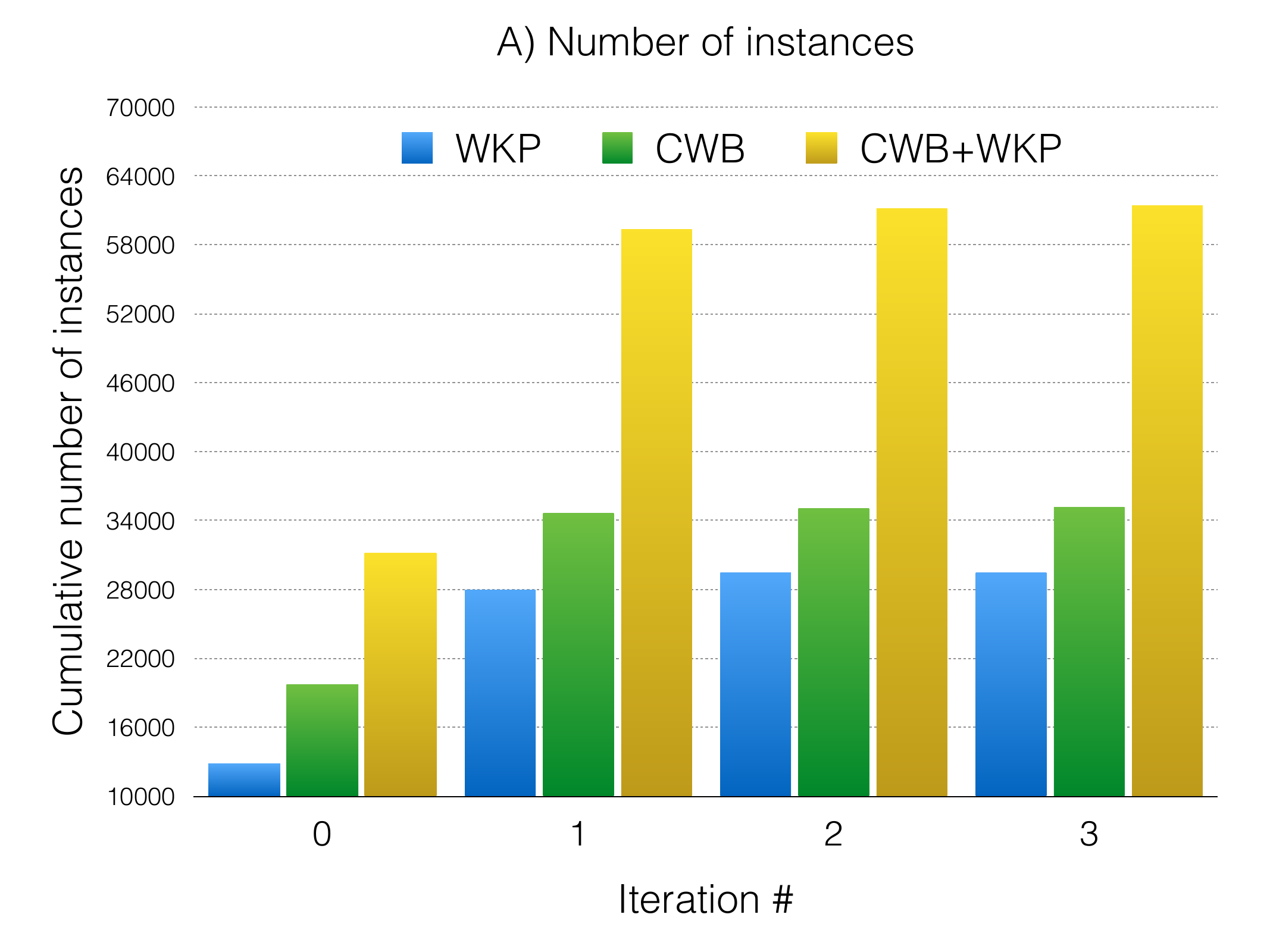}
  \vspace*{-0.4cm}
  \caption{Accumulated number  of instances.}
  \label{fig:instances}
  \end{figure}  
%

%

\section{Conclusion}
In this paper our goal is to address the  bottleneck that has throttled research in higher n-ary relation extraction. To this end, we generated a training data resource for ternary relation extractors.  We described a method for learning and populating ternary relations initially only  using prepositional phrase based templates. Additionally, our method  also learns templates that are not based on prepositions, in an iterative manner. We hope this resource  encourages  research on ternary relation extraction. 

\clearpage
\nocite{*}
\bibliographystyle{acl}
\bibliography{ppternary}

\end{document}